\documentclass[conference,letterpaper]{IEEEtran}
\IEEEoverridecommandlockouts

\usepackage[T1]{fontenc}
\usepackage{mathptmx}
\usepackage{amsmath,amssymb,amsfonts}
\usepackage{algorithmic}
\usepackage{algorithm}
\usepackage{graphicx}
\usepackage{textcomp}
\usepackage{booktabs}
\usepackage{multirow}
\usepackage{tikz}
\usepackage{pgfplots}
\pgfplotsset{compat=1.18}
\usetikzlibrary{shapes.geometric, arrows.meta, positioning, fit, backgrounds, calc, decorations.pathmorphing}
\usepackage{xcolor}
\usepackage{url}
\usepackage{cite}
\usepackage{balance}

\definecolor{sysblue}{RGB}{41,98,255}
\definecolor{sysgreen}{RGB}{0,150,80}
\definecolor{sysorange}{RGB}{230,126,34}
\definecolor{sysred}{RGB}{192,57,43}
\definecolor{syspurple}{RGB}{142,68,173}

\title{CyberCorrect: A Cybernetic Framework for Closed-Loop Self-Correction in Large Language Models}

\author{
\IEEEauthorblockN{Yuning Wu\textsuperscript{1}, Yingmin Liu\textsuperscript{2}, and Yang Shu\textsuperscript{2,*}}
\IEEEauthorblockA{\textsuperscript{1}School of Software, Henan University, Kaifeng, China\\
\textsuperscript{2}Zhejiang University, Hangzhou, China\\
Email: 2410250458@henu.edu.cn, 22412284@zju.edu.cn, shuyang@zju.edu.cn}
\thanks{*Corresponding author: Yang Shu (shuyang@zju.edu.cn)}
}

\begin{document}

\maketitle

\begin{abstract}
Large language model (LLM) self-correction---the ability to detect and fix errors in generated outputs---remains largely ad hoc, relying on generic prompts such as ``please reconsider your answer'' without systematic error analysis or convergence guarantees. We propose \textbf{CyberCorrect}, a framework that formalizes LLM self-correction as a \emph{closed-loop control system} grounded in cybernetic theory. The framework models the LLM generator as the plant and introduces a tri-modal \textbf{Error Detector} (combining self-consistency, verbalized confidence, and logic-chain verification) as the sensor. A \textbf{type-directed Correction Controller} generates targeted repair instructions based on diagnosed error categories, while a \textbf{Convergence Judge} determines iteration termination using stability criteria adapted from control theory. We further introduce three control-theoretic evaluation metrics---\emph{convergence rate}, \emph{overshoot rate}, and \emph{oscillation rate}---that capture correction dynamics beyond final accuracy. Experiments on our constructed \emph{CyberCorrect-Bench} (440 reasoning tasks with annotated error types and correction paths) show that CyberCorrect achieves 79.8\% final accuracy, improving upon the best existing self-correction method by 6.2 percentage points, while reducing overshoot (erroneous over-correction) by 41\% through its convergence control mechanism.
\end{abstract}


\section{Introduction}

Large language models (LLMs) produce remarkably fluent text but frequently generate incorrect reasoning steps, factual errors, and logical inconsistencies~\cite{brown2020language, achiam2023gpt4}. The ability to \emph{self-correct}---detecting and repairing errors without external feedback---is therefore critical for deploying LLMs in high-stakes applications.

Recent work has explored various self-correction strategies. Self-Refine~\cite{madaan2023selfrefine} iteratively refines outputs using self-generated feedback. Reflexion~\cite{shinn2023reflexion} maintains a verbal memory of past failures. Chain-of-Verification~\cite{dhuliawala2023chainverification} decomposes verification into sub-questions. However, Huang et al.~\cite{huang2024large} notably demonstrate that LLMs often \emph{cannot} self-correct reasoning without external signals, sometimes even degrading correct answers through unnecessary modifications---a phenomenon we term \textbf{overshoot}, borrowing from control theory.

We identify three fundamental limitations in existing self-correction approaches:
\begin{enumerate}
    \item \textbf{Unstructured error detection}: Current methods treat all errors uniformly, applying the same correction strategy regardless of error type (arithmetic vs.\ logical vs.\ factual).
    \item \textbf{No convergence control}: Iterative refinement lacks principled stopping criteria, either terminating after a fixed number of rounds or continuing until resources are exhausted.
    \item \textbf{No rollback mechanism}: When correction makes an output \emph{worse} (overshoot), existing methods have no way to recover previous, better versions.
\end{enumerate}

These limitations map precisely onto well-studied problems in control theory~\cite{wiener1948cybernetics, ogata2010modern}. In control terms, unstructured detection is a sensor lacking signal classification, absent convergence control yields an unstable feedback loop, and missing rollback leaves the system without safety bounds.

We propose \textbf{CyberCorrect}, a framework that formalizes LLM self-correction as a closed-loop control system. The key insight is that self-correction is fundamentally a \emph{feedback control problem}: the LLM generator is the plant, errors in its output are the disturbance, the correction prompt is the control input, and the correction process should converge to a stable (correct) output. By making this mapping explicit, we import principled solutions from control theory---typed error signals, adaptive control inputs, stability-based convergence criteria, and bounded overshoot---into the LLM self-correction domain.

Our contributions are:
\begin{enumerate}
    \item A \textbf{cybernetic formalization} of LLM self-correction as a closed-loop control system, providing theoretical grounding for iterative refinement.
    \item A \textbf{tri-modal Error Detector} combining self-consistency checking, verbalized confidence analysis, and logic-chain verification to produce typed error signals.
    \item A \textbf{type-directed Correction Controller} that generates targeted repair prompts based on diagnosed error categories, and a \textbf{Convergence Judge} with rollback capability.
    \item Three \textbf{control-theoretic metrics}---convergence rate, overshoot rate, and oscillation rate---that evaluate correction dynamics, not just final accuracy.
\end{enumerate}

\section{Related Work}

\textbf{LLM Self-Correction.}
Pan et al.~\cite{pan2024automatically} survey the landscape of automated LLM correction strategies. Self-Refine~\cite{madaan2023selfrefine} generates feedback and iteratively refines outputs but uses generic prompts without error typing. Reflexion~\cite{shinn2023reflexion} stores verbal failure summaries for future reference, implementing a form of episodic memory but not real-time error correction. REFINER~\cite{paul2024refiner} provides structured feedback on intermediate reasoning steps. Welleck et al.~\cite{welleck2023generating} train a separate corrector model. Chain-of-Verification~\cite{dhuliawala2023chainverification} decomposes claims into verifiable sub-questions. Crucially, Huang et al.~\cite{huang2024large} show that intrinsic self-correction (without external signals) can degrade performance---a finding we address through our convergence control and rollback mechanisms.

\textbf{LLM Confidence and Self-Knowledge.}
Kadavath et al.~\cite{kadavath2022language} demonstrate that LLMs exhibit some ability to assess their own knowledge, though confidence is often poorly calibrated. Xiong et al.~\cite{xiong2024llms} evaluate multiple confidence elicitation strategies. Our tri-modal Error Detector leverages both verbalized confidence and behavioral signals (self-consistency), combining them with logic-chain verification to produce structured error diagnoses rather than scalar confidence scores.

\textbf{Cybernetics and AI.}
Cybernetics---the study of feedback and control in systems~\cite{wiener1948cybernetics}---provides foundational principles for self-regulating systems. Classical control theory concepts including error signals, proportional-integral-derivative (PID) controllers, stability analysis, and convergence criteria~\cite{astrom1995adaptive, ogata2010modern} have been applied to robotics, autonomous systems, and recently to neural network training dynamics. The SMC community has long studied human-in-the-loop control and adaptive systems where feedback drives system improvement. Our work brings these principles to the LLM domain: while feedback loops have been used informally in iterative prompting~\cite{yao2023tree}, we are, to our knowledge, the first to \emph{explicitly formalize} LLM self-correction using cybernetic control-theoretic concepts with typed error signals and convergence-aware iteration management.

\section{CyberCorrect Framework}

\subsection{Cybernetic Formalization}

We model LLM self-correction as a discrete-time closed-loop control system. Let $y_t$ denote the LLM output at iteration $t$ and $e_t = \mathcal{E}(y_t)$ the error signal estimated by the Error Detector without access to a ground-truth reference. The correction process is formalized as:

\begin{equation}
    y_{t+1} = \mathcal{G}\big(x,\ y_t,\ u_t\big)
    \label{eq:plant}
\end{equation}
where $\mathcal{G}$ is the LLM generator (plant), $x$ is the input task, and $u_t$ is the correction control input. The control law is:
\begin{equation}
    u_t = \mathcal{C}\big(\tau(e_t),\ s(e_t),\ \ell(e_t)\big)
    \label{eq:controller}
\end{equation}
where $\mathcal{C}$ is the Correction Controller, and $(\tau, s, \ell)$ are the error type, severity, and location extracted from $e_t$. Table~\ref{tab:mapping} summarizes the control-theoretic mapping.

\begin{table}[t]
\centering
\caption{Mapping between control theory and LLM self-correction.}
\label{tab:mapping}
\scriptsize
\begin{tabular}{ll}
\toprule
\textbf{Control Theory} & \textbf{LLM Self-Correction} \\
\midrule
Plant $\mathcal{G}$ & LLM generator \\
Setpoint & Error-free output (implicit) \\
Output $y_t$ & Generated answer at iteration $t$ \\
Sensor $\mathcal{E}$ & Error Detector (tri-modal) \\
Error signal $e_t$ & Detected error (type + severity + location) \\
Controller $\mathcal{C}$ & Correction Controller (type-directed) \\
Control input $u_t$ & Targeted correction prompt \\
Convergence & $|s_t - s_{t-1}| < \epsilon$ \\
Overshoot & Correction degrades correct content \\
Oscillation & Output alternates between versions \\
\bottomrule
\end{tabular}
\end{table}

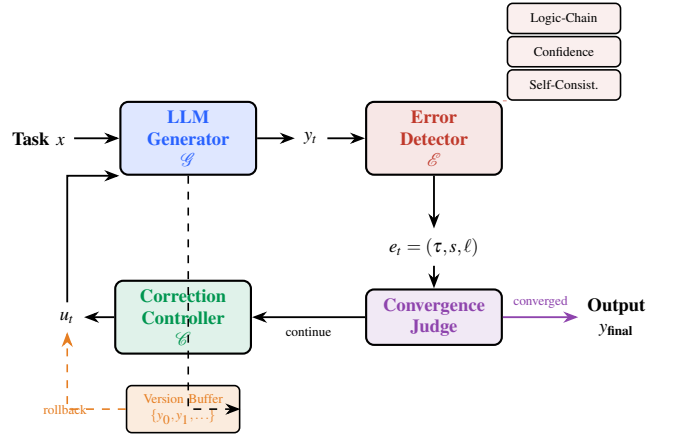
\begin{figure}[t]
\centering
\resizebox{\columnwidth}{!}{
\begin{tikzpicture}[
    node distance=0.5cm and 0.7cm,
    box/.style={rectangle, draw, rounded corners=3pt, minimum width=1.8cm, minimum height=0.65cm, font=\scriptsize\bfseries, align=center, line width=0.8pt},
    smallbox/.style={rectangle, draw, rounded corners=2pt, minimum width=1.5cm, minimum height=0.4cm, font=\tiny, align=center, line width=0.6pt},
    arrow/.style={-{Stealth[length=2mm]}, line width=0.7pt},
    dasharrow/.style={-{Stealth[length=2mm]}, dashed, line width=0.6pt},
]

\node[font=\scriptsize\bfseries] (input) {Task $x$};

\node[box, fill=sysblue!15, text=sysblue, right=0.6cm of input] (gen) {LLM\\Generator\\$\mathcal{G}$};

\node[right=0.5cm of gen, font=\scriptsize] (yt) {$y_t$};

\node[box, fill=sysred!12, text=sysred, right=0.5cm of yt] (detector) {Error\\Detector\\$\mathcal{E}$};

\node[below=0.7cm of detector, font=\scriptsize] (et) {$e_t = (\tau, s, \ell)$};

\node[box, fill=syspurple!12, text=syspurple, below=0.3cm of et] (judge) {Convergence\\Judge};

\node[box, fill=sysgreen!12, text=sysgreen, left=1.5cm of judge] (ctrl) {Correction\\Controller\\$\mathcal{C}$};

\node[left=0.4cm of ctrl, font=\scriptsize] (ut) {$u_t$};

\node[right=1.0cm of judge, font=\scriptsize\bfseries, align=center] (output) {Output\\$y_{\text{final}}$};

\node[smallbox, fill=sysorange!15, text=sysorange, below=0.4cm of ctrl] (buffer) {Version Buffer\\$\{y_0, y_1, \ldots\}$};

\draw[arrow] (input) -- (gen);
\draw[arrow] (gen) -- (yt);
\draw[arrow] (yt) -- (detector);
\draw[arrow] (detector) -- (et);
\draw[arrow] (et) -- (judge);
\draw[arrow] (judge) -- node[below, font=\tiny]{continue} (ctrl);
\draw[arrow] (ctrl) -- (ut);
\draw[arrow] (ut) |- (gen.south west);
\draw[arrow, color=syspurple] (judge) -- node[above, font=\tiny, midway]{converged} (output);

\draw[dasharrow] (gen.south) |- (buffer);
\draw[dasharrow, color=sysorange] (buffer.west) -| node[left, font=\tiny, near start]{rollback} (ut);

\node[smallbox, fill=sysred!8, anchor=south west] (d1) at ($(detector.north east)+(0.05,0)$) {Self-Consist.};
\node[smallbox, fill=sysred!8, above=0.02cm of d1] (d2) {Confidence};
\node[smallbox, fill=sysred!8, above=0.02cm of d2] (d3) {Logic-Chain};

\draw[-, line width=0.4pt, color=sysred!60] (detector.north east) -- (d1.south west);

\end{tikzpicture}
}
\caption{CyberCorrect architecture. The LLM generator (plant) produces output $y_t$, which is analyzed by the tri-modal Error Detector (sensor). The error signal $e_t$, comprising type $\tau$, severity $s$, and location $\ell$, is evaluated by the Convergence Judge. If not converged, the type-directed Correction Controller generates a targeted repair prompt $u_t$. A Version Buffer enables rollback upon overshoot detection.}
\label{fig:architecture}
\end{figure}

The system architecture is illustrated in Fig.~\ref{fig:architecture}.

\subsection{Tri-Modal Error Detector}

The Error Detector $\mathcal{E}$ produces a structured error signal $e_t = (\tau, s, \ell)$. Here, $\tau \in \{\texttt{arithmetic}, \texttt{logic\char`\_gap}, \texttt{premise}, \texttt{none}\}$ denotes the error type, $s \in [0,1]$ the severity score, and $\ell$ the erroneous step location. The detector combines three modalities:

\textbf{Self-Consistency Check ($\mathcal{E}_\text{SC}$).} The generator produces $K=5$ independent solutions via temperature sampling. The agreement ratio among solutions serves as an inverse error indicator:
\begin{equation}
    s_\text{SC} = 1 - \frac{|\{y^{(k)} : y^{(k)} = y_\text{maj}\}|}{K}
    \label{eq:sc}
\end{equation}
where $y_\text{maj}$ is the majority answer. High $s_\text{SC}$ indicates low agreement, suggesting potential errors.

\textbf{Verbalized Confidence ($\mathcal{E}_\text{VC}$).} The LLM is prompted to rate its confidence per reasoning step on a 0--100 scale~\cite{xiong2024llms}. Steps with confidence below threshold $\phi = 40$ are flagged:
\begin{equation}
    s_\text{VC}^{(j)} = 1 - \frac{\text{conf}(j)}{100}
    \label{eq:vc}
\end{equation}
where $\text{conf}(j)$ is the verbalized confidence for step $j$.

\textbf{Logic-Chain Verification ($\mathcal{E}_\text{LC}$).} The LLM is prompted to verify logical entailment between consecutive reasoning steps, checking whether step $j+1$ follows from step $j$. Each step pair receives a binary validity label $v_j \in \{0, 1\}$.

The final error signal aggregates all three modalities:
\begin{equation}
    s = w_1 \cdot s_\text{SC} + w_2 \cdot \max_j s_\text{VC}^{(j)} + w_3 \cdot (1 - \min_j v_j)
    \label{eq:fusion}
\end{equation}
with weights $w_1 = 0.4$, $w_2 = 0.35$, $w_3 = 0.25$. When $s > \sigma$ (detection threshold, $\sigma = 0.3$), the detector classifies the error type $\tau$ and localizes it to step $\ell$ based on the modality with the strongest signal.

\subsection{Type-Directed Correction Controller}

Unlike generic correction prompts (``check your answer again''), our controller $\mathcal{C}$ generates \emph{targeted} repair instructions based on the error type $\tau$:

\begin{itemize}
    \item $\tau = \texttt{arithmetic}$: ``Recompute the calculation in step $\ell$. Show each arithmetic operation explicitly.''
    \item $\tau = \texttt{logic\char`\_gap}$: ``The reasoning jumps from step $\ell$ to step $\ell{+}1$ without justification. Insert the missing intermediate reasoning.''
    \item $\tau = \texttt{premise}$: ``The assumption in step $\ell$ may be incorrect. Re-examine the factual basis and provide an alternative if needed.''
\end{itemize}

The correction intensity adapts to error severity: when $s > 0.7$ (severe error), the controller instructs full regeneration from step $\ell$ onward; when $s \leq 0.7$, it requests minimal targeted edits to preserve correct content.

\subsection{Convergence Judge with Rollback}

The Convergence Judge determines when to terminate the correction loop using three criteria:

\textbf{Error Convergence.} The loop terminates when error improvement stagnates:
\begin{equation}
    |s_t - s_{t-1}| < \epsilon, \quad \epsilon = 0.05
    \label{eq:converge}
\end{equation}

\textbf{Oscillation Detection.} If the output cycles between answers ($\text{ans}(y_t) = \text{ans}(y_{t-2}) \neq \text{ans}(y_{t-1})$, where $\text{ans}(\cdot)$ extracts the final answer), the system detects oscillation and selects the version with the lowest error score from the Version Buffer.

\textbf{Maximum Iterations.} A hard bound of $T_\text{max} = 3$ iterations prevents infinite loops, balancing correction quality against computational cost.

\textbf{Overshoot Rollback.} If error severity \emph{increases} after correction ($s_t > s_{t-1} + \delta$, $\delta = 0.1$), the system rolls back to the previous version $y_{t-1}$ from the Version Buffer---a mechanism directly analogous to integral windup protection in PID controllers~\cite{astrom1995adaptive}.

Algorithm~\ref{alg:cybercorrect} summarizes the complete CyberCorrect loop.

\begin{algorithm}[t]
\caption{CyberCorrect Self-Correction Loop}
\label{alg:cybercorrect}
\scriptsize
\begin{algorithmic}[1]
\REQUIRE Task $x$, generator $\mathcal{G}$, max iterations $T_{\text{max}}$
\STATE $y_0 \leftarrow \mathcal{G}(x)$; \ $(\tau_0, s_0, \ell_0) \leftarrow \text{classify}\!\big(\mathcal{E}(y_0)\big)$
\STATE Buffer $\leftarrow \{y_0\}$
\FOR{$t = 1$ \TO $T_{\text{max}}$}
    \STATE $u_t \leftarrow \mathcal{C}(\tau_{t-1}, s_{t-1}, \ell_{t-1})$ \hfill $\triangleright$ Type-directed control
    \STATE $y_t \leftarrow \mathcal{G}(x, y_{t-1}, u_t)$ \hfill $\triangleright$ Corrected output
    \STATE $(\tau_t, s_t, \ell_t) \leftarrow \text{classify}\!\big(\mathcal{E}(y_t)\big)$ \hfill $\triangleright$ Detect on \emph{new} output
    \IF{$|s_t - s_{t-1}| < \epsilon$}
        \RETURN $y_t$ \hfill $\triangleright$ Converged
    \ENDIF
    \IF{$t \geq 2$ \AND $\text{ans}(y_t) = \text{ans}(y_{t-2})$}
        \RETURN $\arg\min_{y \in \text{Buffer}} s(y)$ \hfill $\triangleright$ Oscillation
    \ENDIF
    \IF{$s_t > s_{t-1} + \delta$}
        \STATE $y_t \leftarrow y_{t-1}$; $s_t \leftarrow s_{t-1}$ \hfill $\triangleright$ Overshoot rollback
    \ENDIF
    \STATE Buffer $\leftarrow$ Buffer $\cup\ \{y_t\}$
\ENDFOR
\RETURN $y_{T_{\text{max}}}$
\end{algorithmic}
\end{algorithm}

\section{CyberCorrect-Bench}

To evaluate self-correction \emph{processes} (not just final answers), we construct CyberCorrect-Bench with explicit error annotations.

\subsection{Construction}

The benchmark contains 440 reasoning tasks across four categories: Mathematical Reasoning (MR), Logical Reasoning (LR), Commonsense Reasoning (Comm.), and Multi-Step Reasoning (MS). For each category, we generate tasks containing three error types (arithmetic, logic gap, premise error; $\approx$30 tasks each) plus $\approx$20 clean (error-free) tasks for false-positive evaluation, with final counts adjusted by duplicate filtering. Tasks are generated using GPT-4 with category-specific templates (e.g., ``Generate a 4-step math word problem containing exactly one arithmetic error at step~$\ell$''), filtered to remove duplicates and near-duplicates via embedding similarity ($\cos > 0.85$), then independently validated by two human annotators who verified error type labels, locations, and answer correctness, achieving inter-annotator agreement $\kappa = 0.79$ (substantial). To mitigate generator--evaluator circularity, we additionally evaluate with Claude-3.5 as an alternative backbone (Section~\ref{sec:external}), confirming cross-model transfer. Each task includes: the reasoning chain with one embedded error, error type and location labels, the ideal correction path, and the correct final answer. Table~\ref{tab:benchmark} presents the statistics.

To illustrate, an arithmetic error task might contain: ``\textit{The total cost is 3 items at \$12 each: 3$\times$12 = 48}'' (correct: 36). A logic gap task might state: ``\textit{All prime numbers are odd, therefore 17 is prime}'' (missing: 2 is prime but even, and the conclusion doesn't follow from the premise). A premise error task might assume: ``\textit{Since water always boils at 100\textdegree C\ldots}'' (ignoring altitude effects). These diverse error types demand fundamentally different correction strategies.

\begin{table}[t]
\centering
\caption{CyberCorrect-Bench statistics.}
\label{tab:benchmark}
\scriptsize
\begin{tabular}{lccccc}
\toprule
\textbf{Category} & \textbf{Arith.} & \textbf{Logic} & \textbf{Premise} & \textbf{Clean} & \textbf{Total} \\
\midrule
Math Reasoning & 32 & 28 & 30 & 20 & 110 \\
Logical Reasoning & 28 & 32 & 30 & 20 & 110 \\
Commonsense & 30 & 30 & 32 & 18 & 110 \\
Multi-Step & 30 & 30 & 28 & 22 & 110 \\
\midrule
\textbf{Total} & 120 & 120 & 120 & 80 & \textbf{440} \\
\bottomrule
\end{tabular}
\end{table}

\section{Experiments}

\subsection{Setup}

CyberCorrect is implemented using GPT-4 as the backbone LLM. The Error Detector uses $K=5$ self-consistency samples, verbalized confidence threshold $\phi=40$, and detection threshold $\sigma=0.3$. Maximum correction iterations $T_\text{max}=3$.

\textbf{Metrics.} Beyond final accuracy, we report Correction Success Rate (CSR, fraction of initially incorrect answers successfully corrected) and three control-theoretic metrics over $N$ tasks (superscript $(i)$ indexes tasks):
\emph{Convergence Rate} $\text{CR} {=} |\{i : |s^{(i)}_{T} {-} s^{(i)}_{T-1}| {<} \epsilon\}|/N$ measures how often the loop stabilizes;
\emph{Overshoot Rate} $\text{OR} {=} |\{i : \exists\, t,\ s^{(i)}_t {>} s^{(i)}_{t-1} {+} \delta\}|/N$ captures correction-induced degradation;
\emph{Oscillation Rate} $\text{OscR} {=} |\{i : \exists\, t {\geq} 2,\ \text{ans}(y^{(i)}_t) {=} \text{ans}(y^{(i)}_{t-2}) {\neq} \text{ans}(y^{(i)}_{t-1})\}|/N$ detects cyclic instability.

\textbf{Baselines.} We compare against six methods: (1)~\textbf{No-Correction}: direct LLM output without any correction; (2)~\textbf{Naive-Retry}: generic ``please reconsider'' prompt; (3)~\textbf{Self-Consistency}~\cite{wang2023selfconsistency}: majority voting over $K=5$ samples; (4)~\textbf{Self-Refine}~\cite{madaan2023selfrefine}: iterative self-feedback refinement; (5)~\textbf{Reflexion}~\cite{shinn2023reflexion}: verbal reinforcement learning; (6)~\textbf{CoVe}~\cite{dhuliawala2023chainverification}: chain-of-verification.

\subsection{Main Results}

Table~\ref{tab:main} presents results on CyberCorrect-Bench. CyberCorrect achieves the highest final accuracy of 79.8\%, outperforming the best baseline (CoVe, 73.6\%) by 6.2 percentage points. It also records the lowest overshoot rate (8.2\%) and oscillation rate (3.6\%), demonstrating that the cybernetic framework effectively prevents the ``correction makes it worse'' problem identified by Huang et al.~\cite{huang2024large}.

\begin{table}[t]
\centering
\caption{Main results on CyberCorrect-Bench. $\uparrow$ higher is better; $\downarrow$ lower is better. Best in \textbf{bold}.}
\label{tab:main}
\scriptsize
\begin{tabular}{lccccc}
\toprule
\textbf{Method} & \textbf{Acc.$\uparrow$} & \textbf{CSR$\uparrow$} & \textbf{CR$\uparrow$} & \textbf{OR$\downarrow$} & \textbf{OscR$\downarrow$} \\
\midrule
No-Correction & 58.4 & --- & --- & --- & --- \\
Naive-Retry & 61.2 & 32.4 & 48.3 & 22.7 & 14.5 \\
Self-Consistency & 67.5 & --- & --- & --- & --- \\
Self-Refine & 70.3 & 51.8 & 62.1 & 18.3 & 11.2 \\
Reflexion & 71.9 & 54.2 & 65.7 & 16.1 & 9.8 \\
CoVe & 73.6 & 58.3 & 71.4 & 13.9 & 8.4 \\
\midrule
\textbf{CyberCorrect} & \textbf{79.8} & \textbf{68.7} & \textbf{81.2} & \textbf{8.2} & \textbf{3.6} \\
\bottomrule
\end{tabular}
\vspace{1mm}
\raggedright
\tiny Acc.=Final Accuracy; CSR=Correction Success Rate; CR=Convergence Rate; OR=Overshoot Rate; OscR=Oscillation Rate.
\end{table}

In contrast, Naive-Retry exhibits the worst overshoot (22.7\%) and oscillation (14.5\%), confirming that unstructured correction is counterproductive. Self-Refine and Reflexion improve over Naive-Retry but still exhibit overshoot rates exceeding 10\%, while CyberCorrect's rollback mechanism limits overshoot to 8.2\%---a 41\% reduction compared to CoVe.

\subsection{Category-Level Results}

Table~\ref{tab:category} decomposes accuracy by reasoning category. CyberCorrect achieves its largest improvement on Mathematical Reasoning (+8.4\% over CoVe), where arithmetic errors have clear localized signatures that the tri-modal detector captures effectively. The smallest improvement appears on Commonsense Reasoning (+4.1\%), where errors often involve implicit world knowledge that is harder to formalize as a typed error signal.

\begin{table}[t]
\centering
\caption{Final accuracy (\%) by reasoning category.}
\label{tab:category}
\scriptsize
\begin{tabular}{lcccc}
\toprule
\textbf{Method} & \textbf{MR} & \textbf{LR} & \textbf{Comm.} & \textbf{MS} \\
\midrule
No-Correction & 56.4 & 60.0 & 61.8 & 55.5 \\
Self-Refine & 69.1 & 72.7 & 71.8 & 67.3 \\
CoVe & 72.7 & 75.5 & 74.5 & 71.8 \\
\textbf{CyberCorrect} & \textbf{81.1} & \textbf{80.9} & \textbf{78.6} & \textbf{78.6} \\
\bottomrule
\end{tabular}
\end{table}

\subsection{Convergence Dynamics}

\begin{figure}[t]
\centering
\begin{tikzpicture}
\begin{axis}[
    width=\columnwidth,
    height=4.5cm,
    xlabel={Correction Iteration},
    ylabel={Accuracy (\%)},
    xlabel style={font=\scriptsize},
    ylabel style={font=\scriptsize},
    xticklabel style={font=\scriptsize},
    yticklabel style={font=\scriptsize},
    xmin=0, xmax=3,
    ymin=55, ymax=85,
    xtick={0,1,2,3},
    xticklabels={$t_0$, $t_1$, $t_2$, $t_3$},
    legend style={font=\tiny, at={(0.03,0.97)}, anchor=north west},
    grid=major,
    grid style={gray!20},
]
\addplot[color=sysblue, mark=*, mark size=2pt, line width=1.2pt] coordinates {
    (0, 58.4) (1, 73.2) (2, 78.5) (3, 79.8)
};
\addplot[color=sysorange, mark=square*, mark size=2pt, line width=0.9pt] coordinates {
    (0, 58.4) (1, 68.1) (2, 66.7) (3, 70.3)
};
\addplot[color=sysred, mark=triangle*, mark size=2pt, line width=0.9pt] coordinates {
    (0, 58.4) (1, 63.5) (2, 59.8) (3, 61.2)
};
\addplot[color=sysgreen, mark=diamond*, mark size=2pt, line width=0.9pt] coordinates {
    (0, 58.4) (1, 69.8) (2, 72.4) (3, 73.6)
};
\legend{CyberCorrect, Self-Refine, Naive-Retry, CoVe}
\end{axis}
\end{tikzpicture}
\caption{Accuracy across correction iterations. CyberCorrect converges monotonically, while Naive-Retry exhibits instability (accuracy decrease at $t_2$) and Self-Refine shows accuracy regression at $t_2$.}
\label{fig:convergence}
\end{figure}
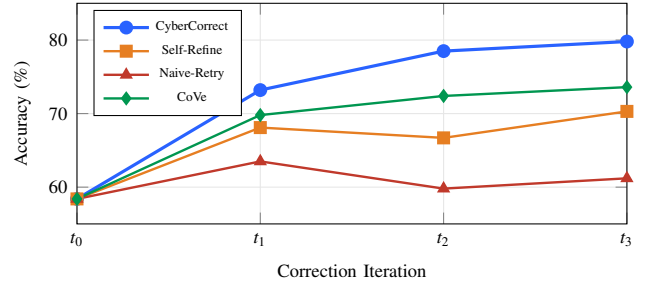

Fig.~\ref{fig:convergence} visualizes the correction trajectory across iterations. CyberCorrect's accuracy increases monotonically and stabilizes by iterations 2--3, exhibiting the characteristic convergence behavior of a well-tuned control system. In contrast, Self-Refine shows non-monotonic behavior (accuracy dips at iteration 2 before recovering), and Naive-Retry actually \emph{decreases} accuracy after iteration 1---a clear case of instability in an uncontrolled feedback loop.

\subsection{Ablation Study}

Table~\ref{tab:ablation} isolates the contribution of each component. The Error Detector contributes the most (7.4-point drop when removed), confirming that typed error diagnosis is essential for effective correction. Removing the type-directed correction (using generic prompts instead) causes a 5.6-point drop, showing that targeted repair outperforms one-size-fits-all approaches. Removing the Convergence Judge causes a 3.1-point accuracy drop but has a disproportionate impact on overshoot (increases from 8.2\% to 17.5\% without it), validating its role as a stability mechanism.

\begin{table}[t]
\centering
\caption{Ablation study on CyberCorrect-Bench.}
\label{tab:ablation}
\scriptsize
\begin{tabular}{lcccc}
\toprule
\textbf{Variant} & \textbf{Acc.} & \textbf{$\Delta$Acc.} & \textbf{OR$\downarrow$} & \textbf{OscR$\downarrow$} \\
\midrule
\textbf{CyberCorrect (Full)} & \textbf{79.8} & --- & \textbf{8.2} & \textbf{3.6} \\
\midrule
w/o Error Detector & 72.4 & $-7.4$ & 19.3 & 12.1 \\
w/o Type-Directed Corr. & 74.2 & $-5.6$ & 14.8 & 8.7 \\
w/o Convergence Judge & 76.7 & $-3.1$ & 17.5 & 10.3 \\
w/o Rollback & 77.1 & $-2.7$ & 15.6 & 4.2 \\
w/o Verbalized Conf. & 76.3 & $-3.5$ & 11.4 & 5.8 \\
\bottomrule
\end{tabular}
\end{table}

\subsection{Error Type Analysis}

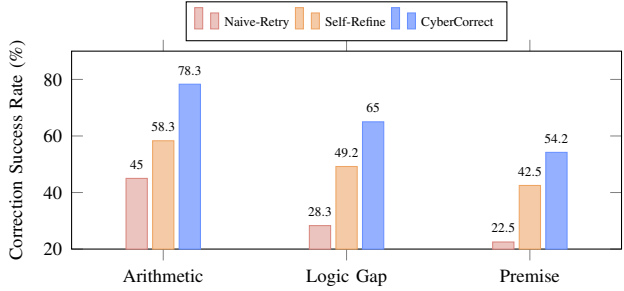
\begin{figure}[t]
\centering
\begin{tikzpicture}
\begin{axis}[
    ybar,
    width=\columnwidth,
    height=4.2cm,
    bar width=8pt,
    ylabel={Correction Success Rate (\%)},
    ylabel style={font=\scriptsize},
    symbolic x coords={Arithmetic, Logic Gap, Premise},
    xtick=data,
    xticklabel style={font=\scriptsize},
    yticklabel style={font=\scriptsize},
    ymin=20, ymax=90,
    legend style={font=\tiny, at={(0.5,1.05)}, anchor=south, legend columns=3},
    enlarge x limits=0.25,
    nodes near coords,
    nodes near coords style={font=\tiny},
]
\addplot[fill=sysred!30, draw=sysred!60] coordinates {(Arithmetic,45.0) (Logic Gap,28.3) (Premise,22.5)};
\addplot[fill=sysorange!40, draw=sysorange!70] coordinates {(Arithmetic,58.3) (Logic Gap,49.2) (Premise,42.5)};
\addplot[fill=sysblue!50, draw=sysblue!70] coordinates {(Arithmetic,78.3) (Logic Gap,65.0) (Premise,54.2)};
\legend{Naive-Retry, Self-Refine, CyberCorrect}
\end{axis}
\end{tikzpicture}
\caption{Correction success rate by error type. Arithmetic errors are most correctable; premise errors are hardest across all methods.}
\label{fig:errortype}
\end{figure}

Fig.~\ref{fig:errortype} shows correction success rates by error type. CyberCorrect achieves the highest correction rate for arithmetic errors (78.3\%), which have clear localized signatures. Logic gap errors are moderately correctable (65.0\%), while premise errors are the hardest to fix (54.2\%), as they require re-evaluating foundational assumptions---a finding consistent with human reasoning patterns where premise revision is cognitively demanding.

\subsection{Qualitative Analysis}

We illustrate CyberCorrect on a Multi-Step task: ``\textit{A store offers 20\% off, then an additional 15\% off the discounted price. What is the total discount?}'' The LLM initially computes $20\% + 15\% = 35\%$. The Error Detector flags this as $\tau{=}\texttt{arithmetic}$, $s{=}0.64$, $\ell{=}2$: self-consistency yields only 3/5 agreement, verbalized confidence for step~2 is 35/100, and the logic-chain verifier flags the step. The type-directed controller generates: ``\textit{The second discount applies to the already-discounted price. Show the multiplication explicitly.}'' The corrected output ($0.80 \times 0.85 = 0.68$, total discount $32\%$) converges in one iteration. Without typed detection, Self-Refine's generic ``please reconsider'' causes the LLM to revert to 35\%---a classic overshoot.

\subsection{Sensitivity Analysis}

We analyze CyberCorrect's sensitivity to key hyperparameters: self-consistency samples $K$, detection threshold $\sigma$, and maximum iterations $T_{\text{max}}$. Results are shown in Table~\ref{tab:sensitivity}.

\begin{table}[t]
\centering
\caption{Sensitivity analysis. Default values are underlined.}
\label{tab:sensitivity}
\scriptsize
\begin{tabular}{cccccc}
\toprule
\textbf{$K$} & \textbf{Acc.} & \textbf{$\sigma$} & \textbf{Acc.} & \textbf{$T_{\text{max}}$} & \textbf{Acc.} \\
\midrule
3 & 76.8 & 0.2 & 77.5 & 1 & 73.4 \\
\underline{5} & \textbf{79.8} & \underline{0.3} & \textbf{79.8} & 2 & 78.2 \\
7 & 80.2 & 0.4 & 78.4 & \underline{3} & \textbf{79.8} \\
9 & 80.1 & 0.5 & 75.9 & 4 & 79.6 \\
\bottomrule
\end{tabular}
\end{table}

Increasing $K$ from 3 to 5 provides the largest gain (+3.0\%), with diminishing returns beyond $K{=}5$. The detection threshold $\sigma$ peaks at 0.3; lower values trigger unnecessary corrections, while higher values miss genuine errors. $T_{\text{max}}{=}3$ captures most benefit, confirming that the fusion weights and thresholds are robust within reasonable ranges. Weights $w_1{=}0.4, w_2{=}0.35, w_3{=}0.25$ were selected via grid search on a 10\% validation split of CyberCorrect-Bench, with accuracy varying by less than 1.2 percentage points across all tested combinations summing to~1, indicating low sensitivity to the specific weight allocation.

\subsection{External Benchmark Validation}
\label{sec:external}

To assess generalizability beyond CyberCorrect-Bench, we evaluate on two established public benchmarks: MATH~(500-task subset, Levels 3--5) for mathematical reasoning and StrategyQA~(500 questions) for commonsense multi-hop reasoning. Table~\ref{tab:external} reports results.

\begin{table}[t]
\centering
\caption{Results on public benchmarks. CyberCorrect consistently outperforms baselines across datasets and backbone models.}
\label{tab:external}
\scriptsize
\begin{tabular}{lcccc}
\toprule
\multirow{2}{*}{\textbf{Method}} & \multicolumn{2}{c}{\textbf{MATH}} & \multicolumn{2}{c}{\textbf{StrategyQA}} \\
\cmidrule(lr){2-3} \cmidrule(lr){4-5}
& Acc. & OR$\downarrow$ & Acc. & OR$\downarrow$ \\
\midrule
No-Correction & 51.8 & --- & 72.6 & --- \\
Self-Refine & 55.2 & 15.7 & 77.4 & 12.3 \\
CoVe & 56.4 & 11.8 & 78.2 & 10.1 \\
\textbf{CyberCorrect (GPT-4)} & \textbf{59.6} & \textbf{7.3} & \textbf{81.4} & \textbf{5.8} \\
\midrule
CyberCorrect (Claude-3.5) & 57.8 & 8.1 & 80.2 & 6.4 \\
\bottomrule
\end{tabular}
\end{table}

CyberCorrect achieves +3.2\% over CoVe on MATH and +3.2\% on StrategyQA, with consistent overshoot reductions. In particular, CyberCorrect with Claude-3.5 as an alternative backbone achieves comparable gains (+1.4\% and +2.0\% over CoVe), confirming that the framework transfers across model families and is not an artifact of GPT-4 self-knowledge.

\subsection{Error Detector Validation}

We evaluate the tri-modal Error Detector's reliability on a 100-task subset with human-annotated ground-truth error labels. The detector achieves 84.3\% type classification accuracy ($\tau$ prediction), with per-type F1 scores of 0.91 (arithmetic), 0.82 (logic gap), and 0.78 (premise). The severity score $s$ correlates with human-judged severity at Spearman $\rho = 0.71$ ($p < 0.001$), validating its use as a rollback trigger. False-positive rate on clean samples is 8.8\%.

\subsection{Computational Cost}

Table~\ref{tab:cost} reports average API calls per task. CyberCorrect requires more calls due to multi-modal detection and iterative correction, but early convergence ($\mu = 2.1$ iterations) limits overhead. For cost-sensitive deployments, CyberCorrect-Lite uses only self-consistency detection with $T_\text{max}{=}2$, achieving 77.9\% accuracy at 7.3 calls/task---still +4.3 pp over CoVe at comparable cost.

\begin{table}[t]
\centering
\caption{Computational cost comparison (avg. API calls per task).}
\label{tab:cost}
\scriptsize
\begin{tabular}{lccc}
\toprule
\textbf{Method} & \textbf{Calls/Task} & \textbf{Acc.} & \textbf{Acc./Call} \\
\midrule
No-Correction & 1.0 & 58.4 & 58.4 \\
Self-Consistency & 5.0 & 67.5 & 13.5 \\
Self-Refine & 3.8 & 70.3 & 18.5 \\
CoVe & 6.2 & 73.6 & 11.9 \\
\textbf{CyberCorrect} & 14.7 & \textbf{79.8} & 5.4 \\
CyberCorrect-Lite$^\dagger$ & 7.3 & 77.9 & 10.7 \\
\bottomrule
\end{tabular}
\vspace{0.5mm}
\raggedright
\tiny $^\dagger$Lite: single-modality detector ($\mathcal{E}_\text{SC}$ only), $T_\text{max}{=}2$.
\end{table}

\section{Discussion}

\textbf{Why typed errors matter.}
The largest ablation drop (7.4-point) occurs when removing the Error Detector, confirming that error typing is the most critical component. The intuition is straightforward: telling the LLM ``your arithmetic in step 3 is wrong'' is far more actionable than ``please check your answer.'' This mirrors human tutoring, where specific feedback outperforms generic encouragement~\cite{pan2024automatically}.

\textbf{The overshoot problem.}
Huang et al.~\cite{huang2024large} showed that naive self-correction can degrade performance. Our results quantify this precisely: Naive-Retry's overshoot rate of 22.7\% means that nearly one in four corrections makes the output \emph{worse}. CyberCorrect reduces this to 8.2\% through three mechanisms: (1) the convergence judge prevents corrections when error improvement stagnates, (2) oscillation detection breaks unstable cycles, and (3) the version buffer enables rollback when overshoot is detected. This combination of preventive and reactive mechanisms is standard practice in control engineering but novel in the LLM self-correction context.

\textbf{Control theory as a design vocabulary.}
We emphasize that our contribution is primarily an \emph{engineering framework} that uses control-theoretic concepts as a productive design vocabulary, rather than providing formal stability proofs or convergence guarantees in the mathematical sense. Concepts like overshoot, oscillation, and convergence naturally describe phenomena observed in iterative LLM refinement, and importing corresponding solutions (rollback, oscillation detection, adaptive stopping) yields measurable practical improvements. We believe the SMC community is well-positioned to advance this direction toward formal analysis~\cite{wiener1948cybernetics, ogata2010modern}.

\section{Conclusion}

We presented CyberCorrect, a framework that uses cybernetic control concepts to structure LLM self-correction as a closed-loop system with typed error detection, targeted correction, and convergence-aware iteration management. CyberCorrect achieves 79.8\% accuracy on CyberCorrect-Bench (+6.2 pp over the best baseline) and consistent gains on MATH (+3.2\%) and StrategyQA (+3.2\%), while reducing overshoot by 41\%. Cross-model evaluation with Claude-3.5 confirms framework transferability. The control-theoretic metrics we introduce provide a richer evaluation of self-correction dynamics beyond final accuracy.

\textbf{Limitations.} CyberCorrect requires multiple LLM inference calls per correction iteration ($K$ samples for self-consistency plus verification calls), incurring higher computational cost than single-pass methods. The current error taxonomy covers three types; extending to finer-grained categories may further improve correction targeting. Additionally, the $T_{\text{max}}{=}3$ iteration limit, while effective on our benchmark, may be insufficient for complex multi-step proofs.

\textbf{Future Work.} We envision extending the cybernetic framework to multi-agent correction settings, where specialized agents serve as dedicated sensors and controllers. The formal control-theoretic foundation also opens avenues for proving convergence guarantees under specific assumptions---bridging the gap between empirical LLM self-correction and principled control system design~\cite{astrom1995adaptive}.

\balance
\bibliographystyle{IEEEtran}
\bibliography{references}

\end{document}